\documentclass[letterpaper]{article} 
\usepackage{aaai2026}  
\usepackage{times}  
\usepackage{helvet}  
\usepackage{courier}  
\usepackage[hyphens]{url}  
\usepackage{graphicx} 
\usepackage{natbib}  
\usepackage{caption} 
\usepackage{amsmath}
\usepackage{booktabs}  

\usepackage{multirow}  
\usepackage{amssymb}
\usepackage{enumitem}

\usepackage{xcolor}
\usepackage{colortbl}

\usepackage{cleveref}

\usepackage{algorithm}
\usepackage[noend]{algpseudocode}

\usepackage{amsfonts}
\newcommand{\simi}[2]{\operatorname{sim}\!\bigl(#1,#2\bigr)}
\algnewcommand{\IfThen}[2]{\State \algorithmicif\ #1\ \algorithmicthen\ #2}
\algnewcommand{\ElseThen}[1]{\State \algorithmicelse\ #1}
\algnewcommand{\ForDo}[2]{\State \algorithmicfor\ #1\ \algorithmicdo\ #2}

\usepackage{newfloat}
\usepackage{listings}

\DeclareCaptionStyle{ruled}{labelfont=normalfont,labelsep=colon,strut=off} 
\floatstyle{ruled}
\newfloat{listing}{tb}{lst}{}
\floatname{listing}{Listing}

\newcommand{\movsem}{\texttt{MovSemCL}}
\newcommand{\movsemmse}{\texttt{MovSemCL-MSE}}
\newcommand{\movsemhse}{\texttt{MovSemCL-HSE}}
\newcommand{\movsemcga}{\texttt{MovSemCL-CGA}}
\title{\movsem{}: Movement-Semantics Contrastive Learning for Trajectory Similarity (Extension)}

\author{
    Zhichen Lai\textsuperscript{\rm 1},
    Hua Lu\textsuperscript{\rm 2}\thanks{Corresponding author.},
    Huan Li\textsuperscript{\rm 3,\rm 4},
    Jialiang Li\textsuperscript{\rm 1},
    Christian S.~Jensen\textsuperscript{\rm 2}
}

\affiliations{
    \textsuperscript{\rm 1}Department of People and Technology, Roskilde University, Roskilde, Denmark\\
    \textsuperscript{\rm 2}Department of Computer Science, Aalborg University, Aalborg/Copenhagen, Denmark\\
    \textsuperscript{\rm 3}The State Key Laboratory of Blockchain and Data Security, Zhejiang University, Hangzhou, China\\
    \textsuperscript{\rm 4}Hangzhou High-Tech Zone (Binjiang) Institute of Blockchain and Data Security, Hangzhou, China\\[2mm]
    zhichenl@ruc.dk,
    luhua@cs.aau.dk,
    lihuancs@zju.edu.cn,
    jiali@ruc.dk,
    csj@cs.aau.dk
}

\begin{document}

\maketitle

\begin{abstract}
Trajectory similarity computation is fundamental functionality that is used for, e.g., clustering, prediction, and anomaly detection. However, existing learning-based methods exhibit three key limitations: (1) insufficient modeling of trajectory semantics and hierarchy, lacking both movement dynamics extraction and multi-scale structural representation; (2) high computational costs due to point-wise encoding; and (3) use of physically implausible augmentations that distort trajectory semantics. To address these issues, we propose \movsem{}, a movement-semantics contrastive learning framework for trajectory similarity computation. \movsem{} first transforms raw GPS trajectories into movement-semantics features and then segments them into patches. Next, \movsem{} employs intra- and inter-patch attentions to encode local as well as global trajectory patterns, enabling efficient hierarchical representation and reducing computational costs. Moreover, \movsem{} includes a curvature-guided augmentation strategy that preserves informative segments (e.g., turns and intersections) and masks redundant ones, generating physically plausible augmented views.
Experiments on real-world datasets show that \movsem{} is capable of outperforming state-of-the-art methods, achieving mean ranks close to the ideal value of 1 at similarity search tasks and  improvements by up to 20.3\% at heuristic approximation, while reducing inference latency by up to 43.4\%.
\end{abstract}

\begin{links}
    \link{Code}{https://github.com/ryanlaics/MovSemCL}
\end{links}

\section{Introduction}
\label{sec:introduction}

The proliferation of GPS-enabled devices enabled massive collections of vehicle trajectory data, enabling applications such as ride-sharing, logistics, and urban analytics~\cite{zheng2015trajectory}. A fundamental operation underlying these applications is \emph{trajectory similarity computation}, which quantifies the similarity between trajectories. This functionality is key to enabling a variety of trajectory applications, like similarity search~\cite{su2020making}, route recommendation~\cite{chen2020learning}, and mobility prediction~\cite{feng2018deepmove}.

Using rigid geometric similarity measures~\cite{hausdorff1914grundzuge,frechet1906quelques}, traditional trajectory similarity computation methods~\cite{cormode2007string,keogh2005exact,vlachos2002discovering} are computationally expensive and ignore underlying semantics. Recent learning-based methods embed trajectories into latent vector spaces using neural architectures such as RNNs~\cite{liu2016predicting,li2024clear,deng2022efficient}, CNNs~\cite{yao2018trajectory,chang2024revisiting}, and Transformers~\cite{xu2020trajectory,chang2023trajcl}, enabling efficient similarity computation. However, these methods still face three key limitations:

\begin{figure}[t]
\centering
\includegraphics[width=\linewidth]{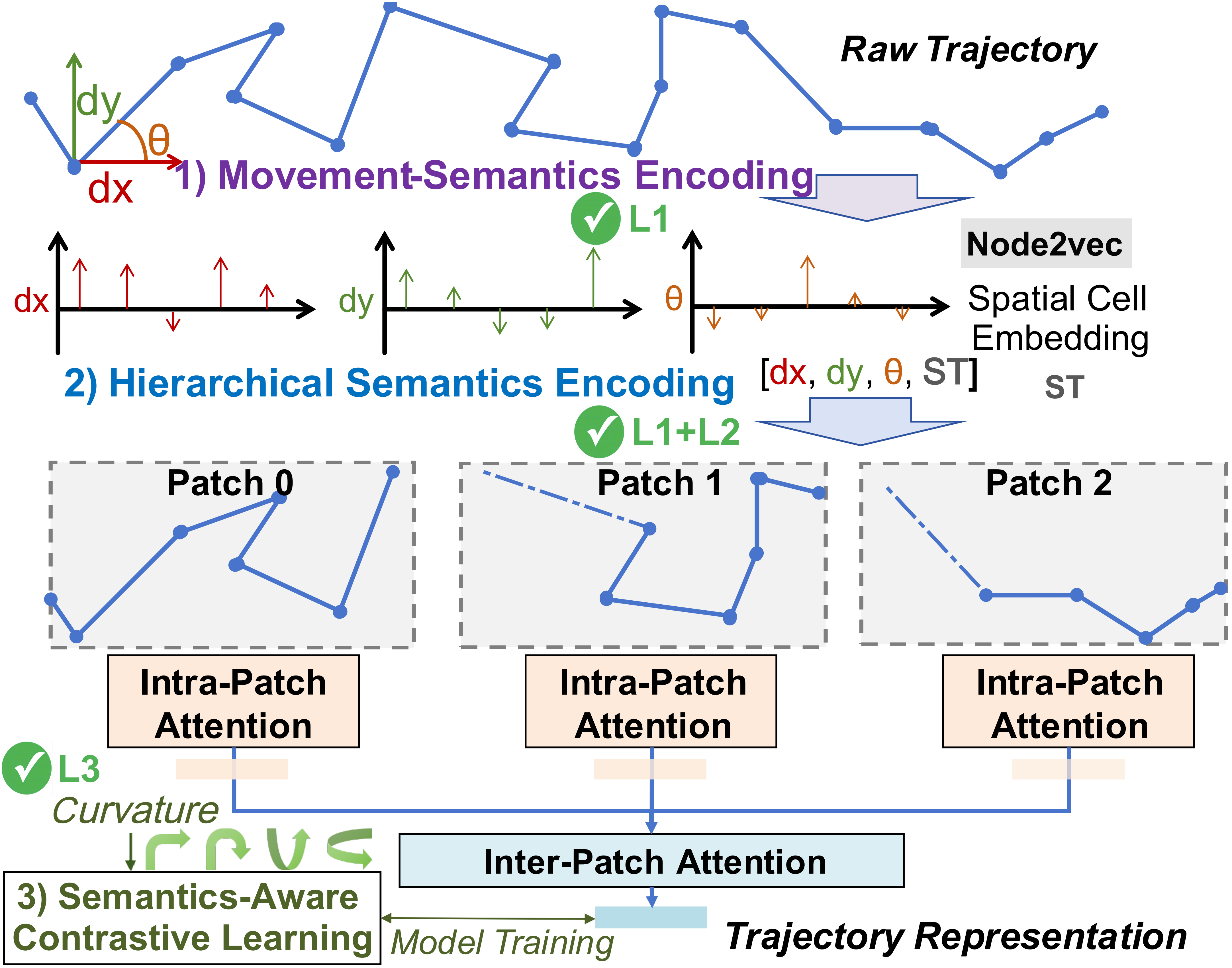}
\caption{Pipeline of \movsem{}. The framework addresses three limitations: Movement-Semantics Encoding extracts movement dynamics (\textbf{L1}), Hierarchical Semantics Encoding captures multi-scale patterns with reduced complexity (\textbf{L1, L2}), and Semantics-Aware Contrastive Learning uses curvature-guided augmentation (\textbf{L3}).}
\label{fig:framework}
\end{figure}
\noindent
\textbf{Limitation 1 (L1): Insufficient modeling of trajectory semantics and hierarchy}. Trajectories contain movement-semantics and are hierarchical in nature—points form maneuvers, maneuvers compose travels—requiring both semantic feature extraction and hierarchical modeling of fine-grained local patterns and coarse-grained global patterns. Existing methods~\cite{liu2016predicting,li2024clear,deng2022efficient,chang2023trajcl,xu2020trajectory} treat trajectories as flat sequences of raw coordinates, failing to capture both movement dynamics and multi-scale structure.

\noindent
\textbf{Limitation 2 (L2): Computational inefficiency for trajectories with many locations}. Real-world trajectories often contain hundreds of points. RNN-based methods~\cite{liu2016predicting,li2024clear,deng2022efficient} lack parallelization, while Transformer-based methods~\cite{xu2020trajectory,chang2023trajcl} scale quadratically with sequence length, forcing lossy downsampling that compromises movement fidelity.

\noindent
\textbf{Limitation 3 (L3): Semantically-unaware augmentations in contrastive learning}. Existing contrastive methods~\cite{chang2023trajcl,li2022cltsim} apply generic augmentations that create physically impossible trajectories—randomly masking points causes spatial jumps, while uniform sampling destroys critical movement patterns like turns and intersections, corrupting signals that are important for learning.

To address these limitations, we introduce \movsem{}, a movement-semantics contrastive learning framework for trajectory similarity computation. As shown in Figure~\ref{fig:framework}, \movsem{} transforms raw GPS sequences into interpretable movement features (see Section~\ref{ssec:feat_extract}), segments them into semantically coherent patches, and encodes both local and global patterns through dual-level attention (see Section~\ref{ssec:hierarchy}). In addition, \movsem{} features a curvature-guided augmentation strategy that generates physically plausible trajectory views by masking redundant segments and preserving behaviorally salient ones (see Section~\ref{ssec:contrastive}).

Our main contributions are as follows:
\begin{itemize}[noitemsep, topsep=0pt]
    \item We propose \movsem{}, a movement-semantics contrastive learning framework for trajectory similarity computation that captures movement dynamics.

    \item We design three key components: movement-semantics encoding captures rich movement semantics (\textbf{L1}), hierarchical patch-based encoding reduces computational complexity from quadratic to near-linear (\textbf{L1, L2}), and curvature-guided augmentation (CGA) generates physically plausible, semantics-aware trajectory augmentations for robust learning (\textbf{L3}).

    \item Extensive experiments on real-world datasets demonstrate that \movsem{} achieves up to \textbf{72.6\%} more accurate similarity search and \textbf{43.4\%} faster inference compared to state-of-the-art baselines.
\end{itemize}

\section{Problem Formulation}
\label{sec:preliminaries}

\paragraph{GPS Trajectory} A GPS trajectory $\mathcal{T}$ is a sequence of timestamped locations:
\begin{equation}
\mathcal{T} =  \langle(s_0, t_0), (s_1, t_1), \cdots, (s_{L-1}, t_{L-1}) \rangle,
\end{equation}
where $s_i = (\mathrm{lon}_i, \mathrm{lat}_i)$ are spatial coordinates, $t_i$ is a timestamp, and $L$ is the trajectory length.

\paragraph{Trajectory Similarity Computation} Given a collection $\mathcal{D} = \{\mathcal{T}_1, \cdots, \mathcal{T}_N\}$ of GPS trajectories with varying lengths, the objective is to learn a representation function $f: \mathcal{D} \rightarrow \mathbb{R}^d$ that maps a trajectory $\mathcal{T} \in \mathcal{D}$ to a fixed-dimensional embedding vector $\mathbf{z} = f(\mathcal{T})$, facilitating efficient similarity computation. The similarity between trajectories $\mathcal{T}_i$ and $\mathcal{T}_j$ is then defined as the Cosine similarity of their embeddings:
\begin{equation}
\text{sim}(\mathcal{T}_i, \mathcal{T}_j) = \frac{\mathbf{z}_i \cdot \mathbf{z}_j}{\|\mathbf{z}_i\| \|\mathbf{z}_j\|}
\end{equation}


\section{Methodology}
\label{sec:framework}

\subsection{Overall Architecture}
As shown in~\Cref{fig:framework}, \movsem{} encompasses three stages:  
(1) \emph{Movement-Semantics Encoding} transforms raw GPS data into movement-semantics features;  
(2) \emph{Hierarchical Semantics Encoding} segments trajectories into patches for hierarchical modeling via dual-level attention;  
(3) \emph{Semantics-Aware Contrastive Learning} leverages physically plausible augmentations and a contrastive loss to focus learning on behaviorally informative segments.

\subsection{Movement-Semantics Encoding}
\label{ssec:feat_extract}

\begin{figure}[ht]
   \centering
   \includegraphics[width=\linewidth]{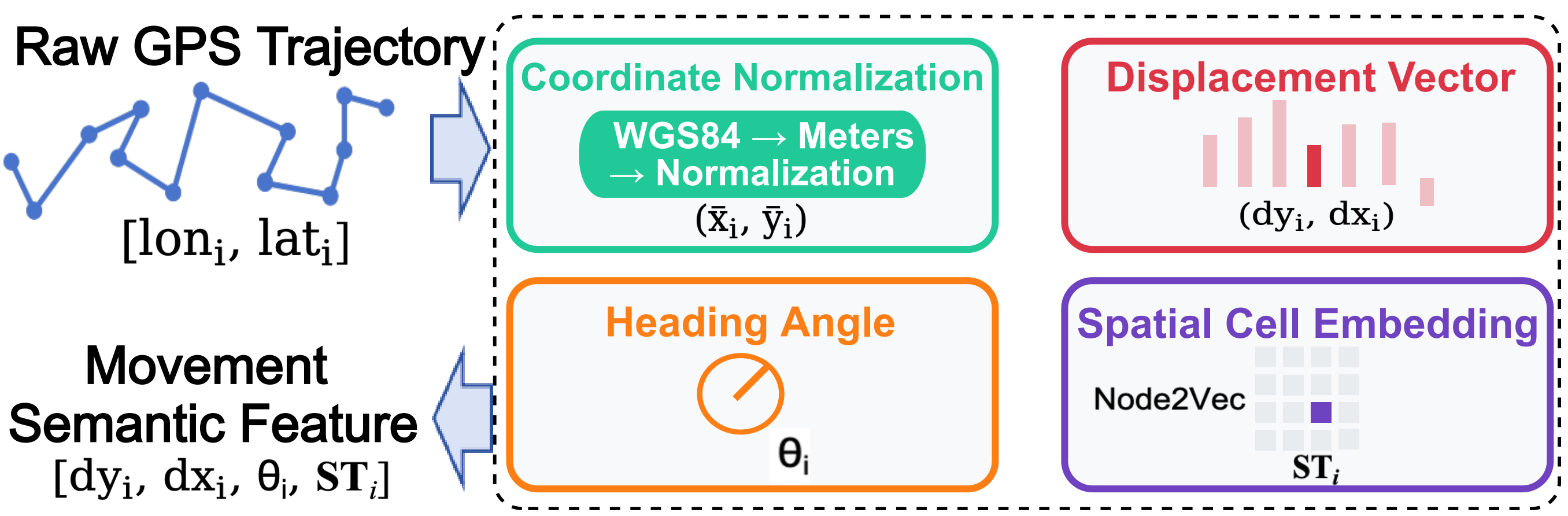}
   \caption{Overview of Movement-Semantics Encoding.}
   \label{fig:movement}
\end{figure}

As illustrated in Figure~\ref{fig:movement}, the proposed Movement-Semantics Encoding facilitates modeling of trajectory semantics by first normalizing GPS coordinates, extracting movement dynamics features (displacement vectors and heading angles), constructing trajectory-induced spatial graphs via Node2Vec~\cite{grover2016node2vec} to encode spatial context into spatial cell embeddings, and composing all features into movement-semantics representations (\textbf{L1}).

\paragraph{Coordinate Normalization} Raw GPS coordinates in the WGS84 coordinate system (longitude, latitude) are not suitable for direct distance computation due to the Earth's curvature. Therefore, we project GPS coordinates to a planar coordinate system using the Mercator projection~\cite{snyder1987map}:
\begin{equation}
(\mathit{mx}_i , \mathit{my}_i) = \mathrm{Mercator}(\mathrm{lon}_i, \mathrm{lat}_i)
\end{equation}

The resulting coordinates are further normalized with respect to the map region of interest:
\begin{equation}
(\bar{x}_i, \bar{y}_i) = \left(\frac{\mathit{mx}_i - x_{\min}^{(m)}}{W_{\mathrm{r}}}, \frac{\mathit{my}_i - y_{\min}^{(m)}}{H_{\mathrm{r}}}\right),
\end{equation}
where $W_{\mathrm{r}}$ and $H_{\mathrm{r}}$ are the width and height of the region.

\paragraph{Movement Dynamics Features} We then compute the displacement vectors and heading angle for each point:
\begin{equation}
dx_i = \begin{cases}
0 & i=0 \\
\bar{x}_{i} - \bar{x}_{i-1} & i>0
\end{cases}, \quad 
dy_i = \begin{cases}
0 & i=0 \\
\bar{y}_{i} - \bar{y}_{i-1} & i>0
\end{cases}.
\end{equation}

Using these displacement vectors, we calculate the heading angle to capture directional changes:
\begin{equation}
\theta_i = \begin{cases}
\frac{\arctan2(dy_i, dx_i)}{\pi} & \text{if } dx_i^2 + dy_i^2 > \epsilon \\
0 & \text{otherwise,}
\end{cases}
\end{equation}
where $\epsilon = 10^{-6}$ prevents division by zero. These features capture directional flow and instantaneous changes, enabling distinction between smooth movements and abrupt maneuvers.

\paragraph{Trajectory-Induced Spatial Graph Construction} To provide spatial context beyond coordinates, we partition the map region into a regular grid of $N_x \times N_y$ cells and assign each trajectory point to a cell according to its normalized coordinates $(\bar{x}_i, \bar{y}_i)$:
\begin{equation}
\label{eq:cell_assignment}
\mathrm{cell}_i = \left\lfloor \frac{\bar{x}_i}{\Delta_x} \right\rfloor \times N_y + \left\lfloor \frac{\bar{y}_i}{\Delta_y} \right\rfloor,
\end{equation}
where $\Delta_x, \Delta_y$ define the grid resolution and $N_y$ is the number of cells along the $y$-axis.

We construct a trajectory-induced directed graph $G = (V, E)$ where each cell is modeled as a node $v \in V$. Given the trajectory collection $\mathcal{D}$, we define edges $e_{ij} \in E$ between cells $i$ and $j$ based on consecutive transitions observed in trajectories:
\begin{equation}
e_{ij} = \begin{cases}
1 & \text{if } \exists \mathcal{T} \in \mathcal{D}, \exists \text{ timestamp } t: \\
 & \quad \text{cell}_t = i \text{ and } \text{cell}_{t+1} = j \\
0 & \text{otherwise}
\end{cases}
\end{equation}

The edge weight $w_{ij}$ represents the number of transitions from cell $i$ to cell $j$:
\begin{equation}
w_{ij} = \sum_{\mathcal{T} \in \mathcal{D}} \sum_{t=0}^{L-2} \mathbf{1}[\text{cell}_{t} = i \text{ and } \text{cell}_{t+1} = j].
\end{equation}

This graph encodes mobility patterns where nodes representing frequently connected cells have large weights, reflecting common movements, while pairs of nodes with no edges represent cells with no immediate movement between them. We apply Node2Vec~\cite{grover2016node2vec} to learn a structural embedding for each cell:
\begin{equation}
\label{eq:cell_embedding}
\mathbf{ST}_i = \mathrm{Node2Vec}(G, \mathrm{cell}_i) \in \mathbb{R}^{d_{\mathrm{se}}}
\end{equation}

\paragraph{Feature Composition} We concatenate all features at each point to form the final movement-semantics representation:
\begin{equation}
\mathbf{f}_i = [dx_i, dy_i, \theta_i, \mathbf{ST}_i] \in \mathbb{R}^{d_{\mathrm{in}}},
\end{equation}
where $d_{\mathrm{in}} = 3 + d_{\mathrm{se}}$. Here, the first three elements encode local movement, while $\mathbf{ST}_i$ captures spatial context.

\subsection{Hierarchical Semantics Encoding}
\label{ssec:hierarchy}

\begin{figure}[htbp]
   \centering
   \includegraphics[width=\linewidth]{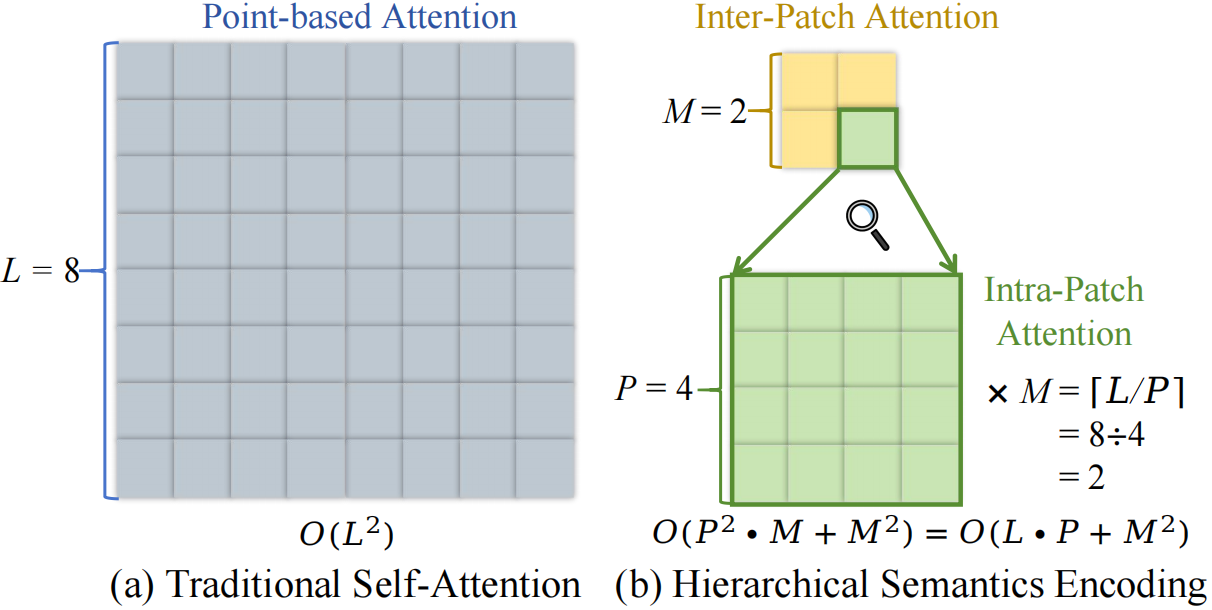}
    \caption{Comparison of attention mechanisms. (a) Traditional Self-Attention with $L=8$. (b) Hierarchical Semantics Encoding with $L=8$, $P=4$, and $M=\lceil L / P \rceil=2$.}
    \label{fig:attention}
\end{figure}

Conventional flat sequence models struggle to capture both local motion details and global intent (\textbf{L1}) and scale poorly to trajectories with many locations (\textbf{L2}). Our hierarchical encoding builds on the movement-semantics features to address \textbf{L1} and \textbf{L2}, enabling efficient modeling of trajectory semantics while reducing computational complexity.

\paragraph{Patch Construction} Given an enriched feature sequence $\langle \mathbf{f}_i\rangle$ from the previous stage, we partition it into $M = \lceil L / P \rceil$ patches, where $P$ is the patch length and $M$ is the number of patches in the trajectory. We obtain the following representation, where a patch $\mathbf{P}_j \in \mathbb{R}^{P \times d_{\mathrm{in}}}$ serves as a locally coherent movement unit:
\begin{equation}
\mathcal{P} =  \langle\mathbf{P}_1, \mathbf{P}_2, \cdots, \mathbf{P}_M \rangle
\end{equation}

As shown in~\Cref{fig:attention}, this patch-based approach reduces computational complexity from $O(L^2)$ in conventional flat sequence models to $O(L \cdot P + M^2)$, where $M = \lceil L / P \rceil$ for typical trajectory lengths, and $M \ll L$ holds, effectively addressing the scalability issues caused by trajectories with many locations (\textbf{L2}). Notably, the dominant term in the complexity depends on the relationship between $L$ and $P$: when $L < P^3$, the $O(L \cdot P)$ term dominates, resulting in near-linear scaling with $L$; otherwise, when $L \geq P^3$, the $O(M^2)$ term becomes dominant, leading to quadratic growth in complexity. Padding and binary masks are used to support variable-length trajectories, following a previous study~\cite{chang2023trajcl}.

\paragraph{Intra-Patch Attention} Within each patch, we employ a self-attention layer to capture patterns unique to each local segment. The intra-patch attention outputs $\mathbf{H}_j^{\mathrm{intra}} \in \mathbb{R}^{P \times d_h}$ are computed as follows.
\begin{equation}
\mathbf{H}_j^{\mathrm{intra}} = \text{SelfAttn}_{\mathrm{intra}}(\mathbf{P}_j + \mathbf{PE}_{\mathrm{local}}),
\end{equation}
where $\mathbf{PE}_{\mathrm{local}}$ provides local positional encoding. To summarize each patch as a fixed-length embedding, we apply masked average pooling over valid (non-padded) positions:
\begin{equation}
\mathbf{h}_j = \frac{1}{\sum_{k=1}^P (1 - m_{j,k}^{\mathrm{intra}})} \sum_{k=1}^P (1 - m_{j,k}^{\mathrm{intra}}) \cdot \mathbf{H}_{j,k}^{\mathrm{intra}},
\end{equation}
where $m_{j,k}^{\mathrm{intra}}$ is a binary mask indicating padding.

\paragraph{Inter-Patch Attention} The patch embedding $\mathbf{H} = [\mathbf{h}_1, \mathbf{h}_2, \cdots, \mathbf{h}_M]^T \in \mathbb{R}^{M \times d_h}$ is then processed by an inter-patch self-attention layer, capturing long-range dependencies and global trajectory intent:
\begin{equation}
\mathbf{H}^{\mathrm{inter}} = \text{SelfAttn}_{\mathrm{inter}}(\mathbf{H} + \mathbf{PE}_{\mathrm{global}}),
\end{equation}
where $\mathbf{PE}_{\mathrm{global}}$ encodes patch-level position information and padded patches are excluded from the computation.

\paragraph{Trajectory Embedding} Finally, we aggregate the outputs of the inter-patch attention to obtain a compact, expressive trajectory embedding:
\begin{equation}
\mathbf{z} = \frac{1}{\sum_{j=1}^M (1 - m_j^{\mathrm{inter}})} \sum_{j=1}^M (1 - m_j^{\mathrm{inter}}) \cdot \mathbf{H}_j^{\mathrm{inter}},
\end{equation}
where $m_j^{\mathrm{inter}}$ masks out padded patches. The resulting embedding $\mathbf{z} \in \mathbb{R}^{d_h}$ robustly preserves both local and global movement semantics of a trajectory.

\subsection{Semantics‑Aware Contrastive Learning}
\label{ssec:contrastive}

\begin{figure}[ht]
    \centering
    \includegraphics[width=\linewidth]{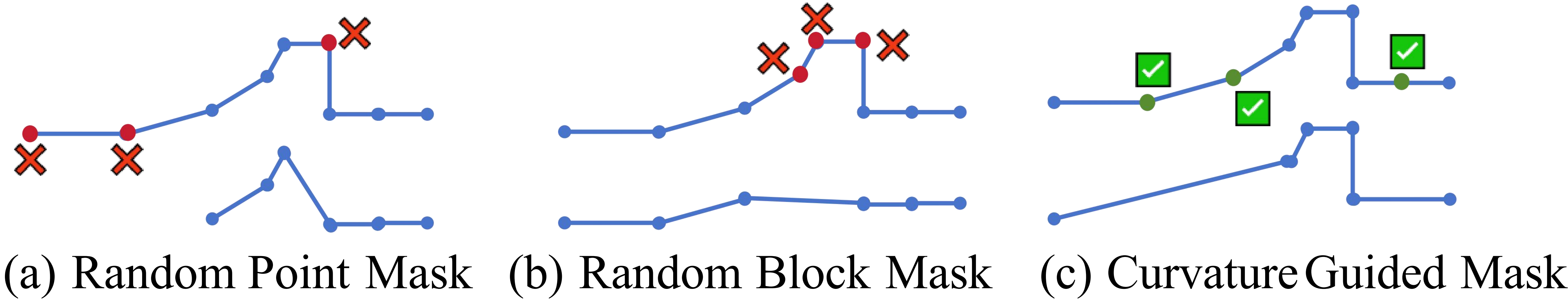}
    \caption{Comparison of trajectory masking strategies.}
    \label{fig:mask}
\end{figure}

While the movement-semantics and hierarchical encodings enable rich trajectory modeling, it is crucial that the model learning focuses on behaviorally informative segments—such as sharp turns or rare maneuvers—rather than overfitting to redundant, straight-line fragments (\textbf{L3}). 

To achieve this, we propose a semantics-aware contrastive learning framework. For each trajectory, we generate \emph{two different augmented views} using \emph{Curvature-Guided Augmentation (CGA)} to create a positive pair, while embeddings of other trajectories in a batch serve as negative samples. This design ensures physically plausible and semantically meaningful augmentations, enhancing the learning by considering behaviorally relevant patterns.

\paragraph{Curvature-Guided Augmentation} Effective contrastive learning hinges on generating semantically consistent and physically plausible augmented views. \movsem{} generates two different augmented views of each trajectory using \emph{Curvature-Guided Augmentation (CGA)}, which form positive pairs for contrastive learning. This augmentation masks trajectory points with probabilities inversely proportional to their local curvature: high-curvature regions (e.g., turns or intersections) are preferentially retained, while straight-line segments are preferentially dropped. Compared to naive random or block masking (Figures~\ref{fig:mask}a--b), CGA produces views that preserve behaviorally informative patterns better, while avoiding spatial discontinuities (Figure~\ref{fig:mask}c). The CGA procedure is detailed in the appendix.

\paragraph{Contrastive Objective} To learn a model, we adopt the MoCo contrastive learning framework~\cite{he2020momentum}, which maintains a dynamic queue of negative embeddings. Given a query embedding~$\mathbf{z}_q$, its corresponding positive key~$\mathbf{z}_k$, a set of negatives~$\{\mathbf{z}_n^{-}\}_{n=1}^N$, and temperature~$\tau$, the contrastive loss is formulated as:
\begin{equation}
\label{eq:infonce}
\mathcal{L}
    = -\frac{\simi{\mathbf{z}_q}{\mathbf{z}_k}}{\tau}
      +\log\!\Bigl(
        e^{\simi{\mathbf{z}_q}{\mathbf{z}_k}/\tau}
        +\sum_{n=1}^{N}e^{\simi{\mathbf{z}_q}{\mathbf{z}_n^{-}}/\tau}
      \Bigr),
\end{equation}
where $\operatorname{sim}(\mathbf{u},\mathbf{v}) = \mathbf{u}^{\!\top}\mathbf{v} / (\|\mathbf{u}\| \|\mathbf{v}\|)$. Only the query encoder is updated via backpropagation; the key encoder is updated using an exponential moving average for stability.

\section{Experiments}
\label{sec:experiments}
To assess the capabilities of \movsem{}, we design experiments to answer the following research questions:
\begin{itemize}[leftmargin=0pt,itemindent=0pt,labelwidth=0pt,labelsep=5pt]
\item \textbf{RQ1 (Effectiveness)}: How does \movsem{} perform at trajectory similarity computation?
\item \textbf{RQ2 (Robustness)}: How robust is \movsem{} under real-world conditions with noisy and degraded trajectories?
\item \textbf{RQ3 (Versatility)}: Can \movsem{} approximate time-consuming heuristic similarity measures with fine-tuning?
\item \textbf{RQ4 (Efficiency)}: What are the computational advantages of \movsem{} for large-scale trajectory processing?
\item \textbf{RQ5 (Component Analysis)}: How do the components of \movsem{} contribute to its performance?
\item \textbf{RQ6 (Hyperparameter Sensitivity)}: How do key hyperparameters affect \movsem{} and what are the best settings?
\end{itemize}

\subsection{Experimental Setup}
\label{subsec:setup}

\paragraph{Datasets} We use two complementary real-world datasets. \textbf{Porto}: dense taxi trajectories from Porto, Portugal (July 2013–June 2014), containing 48 points and being 6.37~km long on average, representing short-distance urban mobility. \textbf{Germany}: sparse long-distance trajectories across Germany (2006–2013), containing 72 points and being 252.49~km long on average, capturing diverse inter-city travel patterns. Following established evaluation protocols~\cite{chang2023trajcl,li2018deep,liu2022cstrm}, we eliminate trajectories with less than 20 or more than 200 points, and we exclude those outside valid geographic regions. For fair comparison with prior work~\cite{chang2023trajcl}, we use a subset of 200 thousand trajectories from the Porto dataset and the full Germany dataset. Each dataset is then split into training/validation/test sets at 70\%/10\%/20\%. From the test set, we reserve 10 thousand samples for the heuristic similarity approximation, partitioned using the same ratio.

\paragraph{Implementation Details} Based on preliminary experiments (see Figure~\ref{fig:hyper}), we set the patch size $P=4$, the embedding dimension $d=256$, the hidden dimension $d_h=256$, and we train for 20 epochs with early stopping. We use the Adam optimizer with a learning rate $1e^{-4}$, batch size 128, and temperature $\tau=0.05$ for contrastive learning. For the spatial discretization, we employ grids with cell sizes adapted to the spatial scale of each dataset: 100 meters for Porto and 1000 meters for Germany. All experiments are conducted on NVIDIA RTX A6000 GPUs. Further details on computational complexity, baselines, and datasets are in the appendix.





\subsection{Trajectory Retrieval Performance (RQ1)}
\label{subsec:retrieval_performance}

We first investigate \movsem{}'s effectiveness on the core task of trajectory similarity computation.

\paragraph{Experimental Protocol} Following baseline methods~\cite{chang2023trajcl}, we evaluate trajectory similarity using a query set $\mathcal{Q}$ and database $\mathcal{D}$ created from 100K test trajectories. We randomly sample 1K trajectories and then split each trajectory $\mathcal{T}^q$ into odd points $\mathcal{T}_a^q = [p_1, p_3, p_5, \cdots]$ and even points $\mathcal{T}_b^q = [p_2, p_4, p_6, \cdots]$. $\mathcal{T}_a^q$ serves as query and $\mathcal{T}_b^q$ as ground-truth in database $\mathcal{D}$, which is augmented with random trajectories to form databases of varying sizes. This splitting creates reasonable similar pairs representing the same movement sequence with different sampling offsets. We report the mean rank of the ground-truth $\mathcal{T}_b^q$ when retrieving similar trajectories for each query $\mathcal{T}_a^q$, with 1 being the ideal rank and smaller ranks being better.
\paragraph{Results and Analysis} Table~\ref{tab:trajectory_similarity} presents the mean rank results across different database sizes. \movsem{} consistently achieves the best performance, with mean ranks very close to 1 across all database sizes.

The performance gap reveals fundamental limitations in existing approaches. Traditional geometric methods (EDR, CSTRM, Hausdorff) measure spatial alignment without movement dynamics, causing severe degradation as geometric similarity becomes ambiguous in large databases. RNN-based methods (t2vec, TrjSR, E2DTC) struggle with long-range dependencies and lack parallelization, while the Transformer-based TrajCL treats trajectories as flat sequences, missing hierarchical movement structure. Additionally, TrajCL's use of random augmentation creates physically implausible trajectories, weakening its contrastive learning. \movsem{}'s near-optimal performance across both dense urban (Porto) and sparse long-distance (Germany) datasets indicates that movement semantics—rather than geometric or temporal patterns alone—provide robust discriminative features that scale effectively. The stability across database sizes reflects a key insight: trajectory similarity is fundamentally about behavioral similarity, which existing methods do not capture comprehensively.

\begin{table}[ht]
\centering
\setlength{\tabcolsep}{2pt} 

\resizebox{\columnwidth}{!}{%
\begin{tabular}{llrrrrr}

\toprule
\textbf{Dataset} & \textbf{Method} & \textbf{20K} & \textbf{40K} & \textbf{60K} & \textbf{80K} & \textbf{100K} \\
\midrule
\multirow{10}{*}{Porto}
 & EDR & 8.318 & 14.398 & 17.983 & 22.902 & 28.753 \\
 & CSTRM & 4.476 & 7.954 & 10.630 & 13.576 & 16.699 \\
 & EDwP & 3.280 & 4.579 & 5.276 & 6.191 & 7.346 \\
 & Hausdorff & 3.068 & 4.014 & 4.649 & 5.451 & 6.376 \\
 & Fréchet & 3.560 & 4.959 & 5.968 & 7.192 & 8.631 \\
 & t2vec & 1.523 & 2.051 & 2.257 & 2.612 & 3.068 \\
 & TrjSR & 1.876 & 2.783 & 3.208 & 3.826 & 4.635 \\
 & E2DTC & 1.560 & 2.111 & 2.349 & 2.731 & 3.213 \\
 & CLEAR & 1.435 & 1.593 & 1.766 & 1.923 & 2.077 \\
 & TrajCL & \underline{1.005} & \underline{1.006} & \underline{1.006} & \underline{1.007} & \underline{1.010} \\
 & \movsem{} & \cellcolor{gray!20}\textbf{1.002} & \cellcolor{gray!20}\textbf{1.004} & \cellcolor{gray!20}\textbf{1.005} & \cellcolor{gray!20}\textbf{1.005} & \cellcolor{gray!20}\textbf{1.005} \\
\midrule
\multirow{10}{*}{Germany}
 & EDR & 279.385 & 558.288 & 834.208 & 1108.975 & 1370.004 \\
 & CSTRM & OOM & OOM & OOM & OOM & OOM \\
 & EDwP & {2.168} & {2.277} & {2.371} & {2.454} & {2.515} \\
 & Hausdorff & 2.803 & 3.509 & 4.206 & 4.906 & 5.551 \\
 & Fréchet & 2.581 & 3.108 & 3.633 & 4.113 & 4.589 \\
 & t2vec & 1.571 & 1.982 & 2.387 & 2.718 & 3.053 \\
 & TrjSR & 6.517 & 11.741 & 16.969 & 22.182 & 24.083 \\
 & E2DTC & 3.136 & 5.156 & 7.248 & 9.207 & 10.956 \\
 & CLEAR & 1.104 & 1.138 & 1.177 & 1.202 & 1.222 \\
 & TrajCL & \underline{1.012} & \underline{1.022} & \underline{1.034} & \underline{1.040} & \underline{1.045} \\
 & \movsem{} & \cellcolor{gray!20}\textbf{1.002} & \cellcolor{gray!20}\textbf{1.003} & \cellcolor{gray!20}\textbf{1.003} & \cellcolor{gray!20}\textbf{1.005} & \cellcolor{gray!20}\textbf{1.008} \\
\bottomrule
\end{tabular}%
}
\caption{Mean rank vs. database size (RQ1). Best results are in \textbf{bold}, second-best are \underline{underlined}. \movsem{} consistently achieves the best mean rank (ideal=1).}
\label{tab:trajectory_similarity}

\end{table}

\subsection{Robustness Evaluation (RQ2)}
\label{subsec:robustness}

We proceed to examine \movsem{}'s robustness to data degradation that occurs commonly in real-world settings.

\paragraph{Down-sampling Robustness} GPS trajectories often suffer from missing data points due to signal loss, battery constraints, or privacy-preserving sampling. Following prior studies~\cite{chang2023trajcl,li2024clear}, we down-sample trajectories in $\mathcal{Q}$ and $\mathcal{D}$ by randomly masking points in each trajectory with a probability $\rho_s \in [0.1, 0.5]$, while keeping $|\mathcal{D}| = 100,000$. Table~\ref{tab:downsample_rank} shows that \movsem{} achieves the best performance across all down-sampling rates. Among deep learning methods, TrajCL performs well at low rates but deteriorates significantly as degradation increases (36.352 at 0.5 on Porto). CLEAR, designed with augmentation strategies for robustness, maintains more stable performance on intra-city dataset Porto but degrades on inter-city trajectories like Germany. However, \movsem{} exhibits the best robustness across most settings, clearly demonstrating that movement-semantics modeling provides inherent robustness to data sparsity.

\paragraph{Distortion Robustness} Real GPS data contains coordinate inaccuracies due to sensor noise and atmospheric interference. We follow prior studies~\cite{chang2023trajcl,li2024clear} and apply random coordinate shifts to a proportion $\rho_d \in [0.1, 0.5]$ of trajectory points. Table~\ref{tab:distort_rank} shows that \movsem{} again exhibits the best robustness, maintaining mean ranks very close to 1 across all distortion rates.

\begin{table}[ht]
\centering
\setlength{\tabcolsep}{2pt} 

\resizebox{\columnwidth}{!}{%
\begin{tabular}{llrrrrr}
\toprule
\textbf{Dataset} & \textbf{Method} & \textbf{0.1} & \textbf{0.2} & \textbf{0.3} & \textbf{0.4} & \textbf{0.5} \\
\midrule
\multirow{10}{*}{Porto}
 & EDR & 57.173 & 203.993 & 806.033 & 2286.821 & 4872.231 \\
 & CSTRM & 24.794 & 47.137 & 123.124 & 257.540 & 687.262 \\
 & EDwP & 8.442 & 10.968 & 18.727 & 28.394 & 68.061 \\
 & Hausdorff & 10.026 & 23.293 & 56.561 & 89.827 & 275.206 \\
 & Fréchet & 10.668 & 18.516 & 29.740 & 93.851 & 181.271 \\
 & t2vec & 4.786 & 8.461 & 19.689 & 35.219 & 115.364 \\
 & TrjSR & 7.941 & 15.746 & 151.948 & 549.108 & 1341.883 \\
 & E2DTC & 5.100 & 9.385 & 21.845 & 39.402 & 124.320 \\
 & CLEAR & 1.518 & 1.914 & 2.792 & \underline{3.650} & \textbf{6.090} \\
 & TrajCL & \underline{1.026} & \underline{1.191} & \textbf{1.513} & {3.847} & {36.352} \\
 & \movsem{} & \cellcolor{gray!20}\textbf{1.018} & \cellcolor{gray!20}\textbf{1.098} & \cellcolor{gray!20}\underline{1.682} & \cellcolor{gray!20}\textbf{1.961} & \cellcolor{gray!20}\underline{9.951} \\
\midrule
\multirow{10}{*}{Germany}
 & EDR & 1368.829 & 1379.489 & 1375.261 & 1380.517 & 1389.433 \\
 & EDwP & 2.173 & 2.509 & 2.176 & 2.191 & 2.209 \\
 & Hausdorff & 2.514 & 2.742 & 4.353 & 4.448 & 5.627 \\
 & Fréchet & 2.358 & 2.492 & 3.735 & 3.824 & 4.642 \\
 & CSTRM & OOM & OOM & OOM & OOM & OOM \\
 & t2vec & 4.453 & 6.736 & 9.087 & 9.470 & 9.775 \\
 & TrjSR & 24.539 & 30.318 & 55.002 & 68.070 & 111.175 \\
 & E2DTC & 11.595 & 13.478 & 15.843 & 18.532 & 19.134 \\
 & CLEAR & 1.265 & 1.276 & 1.396 & 1.460 & \underline{1.740} \\
 & TrajCL & \underline{1.048} & \underline{1.050} & \textbf{1.059} & \underline{1.418} & {2.045} \\
 & \movsem{} & \cellcolor{gray!20}\textbf{1.001} & \cellcolor{gray!20}\textbf{1.008} & \cellcolor{gray!20}\underline{1.080} & \cellcolor{gray!20}\textbf{1.151} & \cellcolor{gray!20}\textbf{1.265} \\
\bottomrule
\end{tabular}%
}
\caption{Mean rank vs. down-sampling rate (RQ2). }
\label{tab:downsample_rank}

\end{table}

\begin{table}[ht]
\centering

\setlength{\tabcolsep}{2pt} 

\resizebox{\columnwidth}{!}{%
\begin{tabular}{llrrrrr}
\toprule
\textbf{Dataset} & \textbf{Method} & \textbf{0.1} & \textbf{0.2} & \textbf{0.3} & \textbf{0.4} & \textbf{0.5} \\
\midrule
\multirow{10}{*}{Porto}
 & EDR & 28.243 & 28.498 & 27.899 & 28.070 & 28.932 \\
 & EDwP & 7.591 & 7.166 & 7.038 & 7.235 & 7.236 \\
 & Hausdorff & 6.549 & 6.737 & 6.706 & 6.592 & 6.739 \\
 & Fréchet & 8.689 & 8.854 & 8.755 & 8.636 & 9.083 \\
 & CSTRM & 20.860 & 20.081 & 22.081 & 24.688 & 26.243 \\
 & t2vec & 3.212 & 3.487 & 3.981 & 3.897 & 3.999 \\
 & TrjSR & 4.781 & 5.087 & 35.144 & 6.194 & 7.201 \\
 & E2DTC & 3.348 & 3.678 & 4.210 & 4.129 & 4.222 \\
 & CLEAR & 1.345 & 1.313 & 1.330 & 1.309 & 1.356 \\
 & TrajCL & \underline{1.022} & \underline{1.154} & \underline{1.076} & \underline{1.091} & \underline{1.039} \\
 & \movsem{} & \cellcolor{gray!20}\textbf{1.004} & \cellcolor{gray!20}\textbf{1.012} & \cellcolor{gray!20}\textbf{1.005} & \cellcolor{gray!20}\textbf{1.006} & \cellcolor{gray!20}\textbf{1.004} \\
\midrule
\multirow{10}{*}{Germany}
 & EDR & 1373.985 & 1372.984 & 1373.981 & 1373.966 & 1373.944 \\
 & EDwP & 2.488 & 2.489 & 2.492 & 2.489 & 2.489 \\
 & Hausdorff & 5.587 & 5.576 & 5.573 & 5.566 & 5.568 \\
 & Fréchet & 4.631 & 4.625 & 4.609 & 4.625 & 4.612 \\
 & CSTRM & OOM & OOM & OOM & OOM & OOM \\
 & t2vec & 3.863 & 3.976 & 4.903 & 3.580 & 3.625 \\
 & TrjSR & 27.146 & 27.156 & 27.032 & 26.935 & 27.035 \\
 & E2DTC & 10.946 & 11.161 & 10.940 & 11.275 & 10.693 \\
 & CLEAR & 1.223 & 1.194 & 1.209 & 1.190 & 1.233 \\
 & TrajCL & \underline{1.049} & \underline{1.051} & \underline{1.049} & \underline{1.062} & \underline{1.054} \\
 & \movsem{} & \cellcolor{gray!20}\textbf{1.008} & \cellcolor{gray!20}\textbf{1.008} & \cellcolor{gray!20}\textbf{1.008} & \cellcolor{gray!20}\textbf{1.008} & \cellcolor{gray!20}\textbf{1.007} \\
\bottomrule
\end{tabular}%
}
\caption{Mean rank vs. distortion rate (RQ2).} 
\label{tab:distort_rank}

\end{table}

\subsection{Heuristic Similarity Approximation (RQ3)}
\label{subsec:heuristic_approximation}

Beyond similarity search, we evaluate whether or not \movsem{}'s learned representations can effectively approximate traditional distance measures, thereby assessing the versatility of its movement-semantics embeddings.

\paragraph{Experimental Protocol} Following the prior work~\cite{chang2023trajcl}, we fine-tune the pre-trained \movsem{} encoder with a two-layer multilayer perceptron head to predict EDR, EDwP, Hausdorff, and Fréchet distances using the MSE loss. This setup tests whether the movement-semantics representations capture sufficient information to approximate these time-consuming heuristic measures. We compare against both fine-tuning models and supervised methods trained specifically for distance prediction, including TrajSimVec~\cite{zhang2020trajectory}, TrajGAT~\cite{yao2022trajgat}, and T3S~\cite{yang2021t3s}. We report HR@$k$ (Hit Ratio at $k$), the proportion of ground-truth top-$k$ similar trajectories correctly identified in predicted top-$k$ results, and R5@20 (Recall of top-5 in top-20), measuring recall of returning ground-truth top-5 trajectories in the top-20 results.

\paragraph{Results} Table~\ref{tab:tablex} shows that \movsem{} achieves an average rank of 1 across all measures. Notably, \movsem{} demonstrates improvements over TrajCL: 20.3\% improvement on EDR, 1.7\% on Hausdorff, and 1.8\% on Fréchet for the HR@5 metrics. The consistently high R5@20 scores ($> 0.95$) for Hausdorff and Fréchet indicate that the movement-semantics representations does very well at capturing the geometric properties these measures emphasize.

\begin{table*}[ht]
\centering
\footnotesize
\setlength{\tabcolsep}{2pt} 
\begin{tabular}{lccccccccccccc}
\toprule
\multirow{2}{*}{\textbf{Method}} & \multicolumn{3}{c}{\textbf{EDR}} & \multicolumn{3}{c}{\textbf{EDwP}} & \multicolumn{3}{c}{\textbf{Hausdorff}} & \multicolumn{3}{c}{\textbf{Fréchet}} & \multirow{2}{*}{\parbox{1.2cm}{\centering\textbf{Average}\\\textbf{rank}}} \\
\cmidrule(lr){2-4} \cmidrule(lr){5-7} \cmidrule(lr){8-10} \cmidrule(lr){11-13}
& \textbf{HR@5} & \textbf{HR@20} & \textbf{R5@20} & \textbf{HR@5} & \textbf{HR@20} & \textbf{R5@20} & \textbf{HR@5} & \textbf{HR@20} & \textbf{R5@20} & \textbf{HR@5} & \textbf{HR@20} & \textbf{R5@20} & \\
\midrule
t2vec & 0.125 & 0.164 & 0.286 & 0.399 & 0.518 & 0.751 & 0.405 & 0.549 & 0.770 & 0.504 & 0.651 & 0.883 & 6 \\
TrjSR & 0.137 & 0.147 & 0.273 & 0.271 & 0.346 & 0.535 & 0.541 & 0.638 & 0.880 & 0.271 & 0.356 & 0.523 & 9 \\
E2DTC & 0.122 & 0.157 & 0.272 & 0.390 & 0.514 & 0.742 & 0.391 & 0.537 & 0.753 & 0.498 & 0.648 & 0.879 & 7 \\
CSTRM & 0.138 & 0.191 & 0.321 & 0.415 & 0.536 & 0.753 & 0.459 & 0.584 & 0.813 & 0.421 & 0.557 & 0.768 & 4 \\
CLEAR & 0.158 & 0.207 & 0.351 & 0.487 & 0.594 & 0.831 & 0.596 & 0.687 & 0.936 & 0.583 & 0.709 & 0.937 & 3 \\
TrajCL & \underline{0.172} & \underline{0.222} & \underline{0.376} & \textbf{0.546} & \textbf{0.646} & \textbf{0.881} &{0.643} & {0.721} & {0.954} & \underline{0.618} & \underline{0.740} & \underline{0.955} & 2 \\
\movsem{} & \cellcolor{gray!20}\textbf{0.207} & \cellcolor{gray!20}\textbf{0.308} & \cellcolor{gray!20}\textbf{0.487} & \cellcolor{gray!20}\underline{0.536} & \cellcolor{gray!20}\underline{0.642} & \cellcolor{gray!20}\underline{0.873} & \cellcolor{gray!20}\underline{0.654} & \cellcolor{gray!20}\textbf{0.742} & \cellcolor{gray!20}\textbf{0.970} & \cellcolor{gray!20}\textbf{0.629} & \cellcolor{gray!20}\textbf{0.741} & \cellcolor{gray!20}\textbf{0.957} & \cellcolor{gray!20}\textbf{1} \\
\cmidrule{1-14}
TrajSimVec & 0.119 & 0.163 & 0.285 & 0.172 & 0.253 & 0.390 & 0.339 & 0.429 & 0.543 & 0.529 & 0.664 & 0.894 & 10 \\
TrajGAT & 0.090 & 0.102 & 0.184 & 0.201 & 0.274 & 0.469 & \textbf{0.686} & \underline{0.740} & \underline{0.969} & 0.362 & 0.403 & 0.704 & 8 \\
T3S & 0.140 & 0.192 & 0.325 & 0.377 & 0.498 & 0.702 & 0.329 & 0.482 & 0.668 & 0.595 & 0.728 & 0.946 & 5 \\
\bottomrule
\end{tabular}
\caption{HR@5, HR@20, and R5@20 of self-supervised and supervised methods to approximate heuristic measures on Porto (RQ3). }
\label{tab:tablex}
\end{table*}

\subsection{Efficiency Analysis (RQ4)}
\label{subsec:efficiency}
To assess deployment feasibility, we compare \movsem{}'s computational efficiency with that of the SOTA method TrajCL, which also adopts self-attention-based encoder.

\paragraph{Inference Efficiency} Table~\ref{tab:sota} compares \movsem{} with TrajCL under identical settings (embedding size = 256, batch size = 128, same hardware). \movsem{} achieves significant improvements across all metrics: 41.2\% fewer FLOPs, 43.4\% faster inference, and 76.6\% higher throughput. These gains arise from replacing the global $O(L^2)$ attention with block-wise computations of $O(L \cdot P + M^2)$.

\begin{figure}[htbp]
    \centering
    \includegraphics[width=\linewidth]{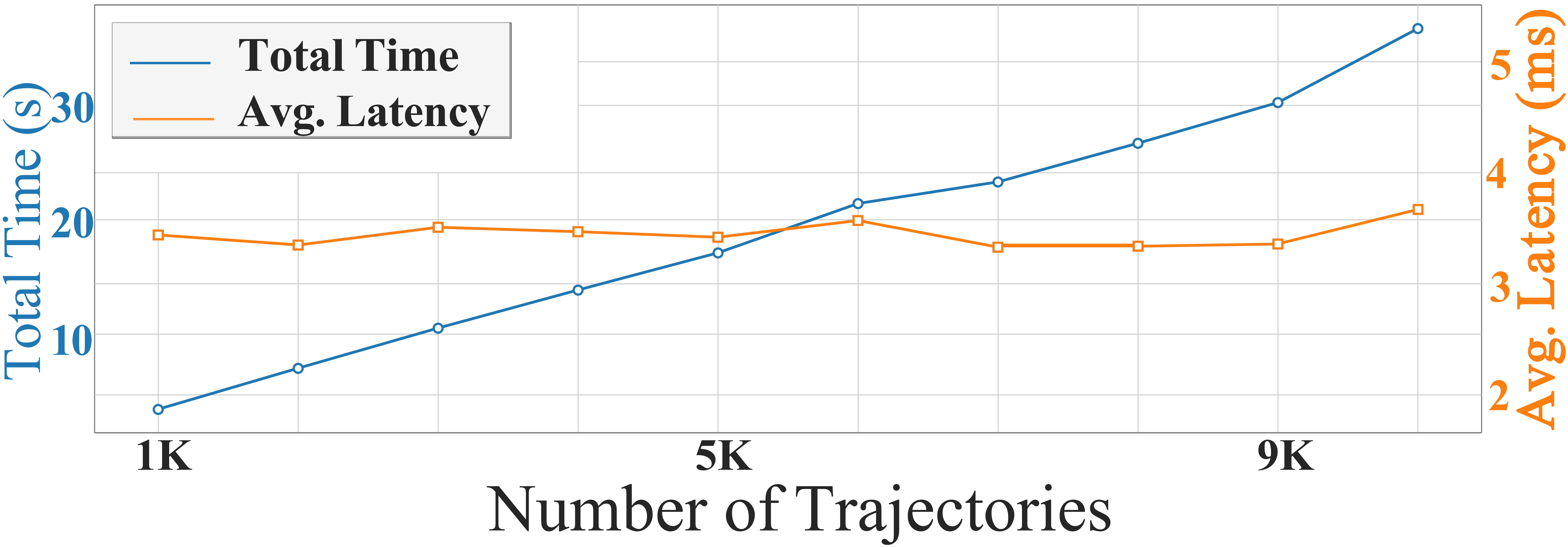}
    \caption{Scalability analysis (RQ4). }
\label{fig:throughput}
\end{figure}

\paragraph{Scalability Analysis} Figure~\ref{fig:throughput} reports \movsem{}'s scalability across varying workloads. The total processing time scales linearly with the number of trajectories, confirming performance without computational bottlenecks. Crucially, the per-sample latency remains remarkably stable (at $\sim$3.4ms) regardless of the dataset size, indicating consistent high efficiency from small batches to large-scale processing.

\begin{table}[ht]
\centering
\footnotesize
\resizebox{\columnwidth}{!}{%

\begin{tabular}{lrrr}
\toprule
\textbf{Metric} & {TrajCL} & \movsem{} & {Improv.} \\
\midrule
FLOPs (M) & 158.69 & \textbf{93.34} & \cellcolor{gray!20}{41.2\%} \\
Latency (ms) & 6.08 & \textbf{3.44} & \cellcolor{gray!20}{43.4\%} \\
Throughput (samples/s) & 164.46 & \textbf{290.41} & \cellcolor{gray!20}{76.6\%} \\
\bottomrule
\end{tabular}}
\caption{Efficiency comparison (RQ4). \movsem{} achieves substantial efficiency improvements over TrajCL.}
\label{tab:sota}

\end{table}

\subsection{Ablation Study (RQ5)}
\label{subsec:ablation}

To evaluate the contribution of each system component, we remove the movement-semantics encoding (MSE, using only cell embedding), the hierarchical semantics encoding (HSE, using flat sequence processing), and the curvature-guided augmentation (CGA, using random point mask). 

\paragraph{Results} Table~\ref{tab:ablation_study} reveals a clear hierarchy: MSE has the most critical impact, with its removal causing severe degradation, offering evidence that movement semantics are essential. CGA provides moderate but consistent improvements, validating the semantics-aware masking. HSE contributes the least but still provides meaningful improvements, confirming that hierarchical encoding is beneficial.

\begin{table}[ht]
\centering
\footnotesize 
\begin{tabular}{lrrrr}
\toprule
\textbf{Method} & \multicolumn{2}{c}{\textbf{Porto}} & \multicolumn{2}{c}{\textbf{Germany}} \\
\cmidrule(lr){2-3} \cmidrule(lr){4-5}
& 20K & 100K & 20K & 100K \\
\midrule
\movsemmse{} & 1.521 & 3.045 & 1.595 & 4.122 \\
\movsemhse{} & 1.005 & 1.012 & 1.010 & 1.039 \\
\movsemcga{} & 1.033 & 1.098 & 1.180 & 1.234 \\
\movsem{} & \cellcolor{gray!20}\textbf{1.002} & \cellcolor{gray!20}\textbf{1.005} & \cellcolor{gray!20}\textbf{1.002} & \cellcolor{gray!20}\textbf{1.008} \\
\bottomrule
\end{tabular}
\caption{Ablation study (RQ5). MSE is the most impactful module, and all modules are effective.}
\label{tab:ablation_study}

\end{table}

\subsection{Hyperparameter Study (RQ6)}
\label{subsec:parameter_studies}

Finally, we examine \movsem{}'s sensitivity to key hyperparameters on Porto to provide implementation guidance.

\begin{figure}[ht]
    \centering
    \includegraphics[width=\linewidth]{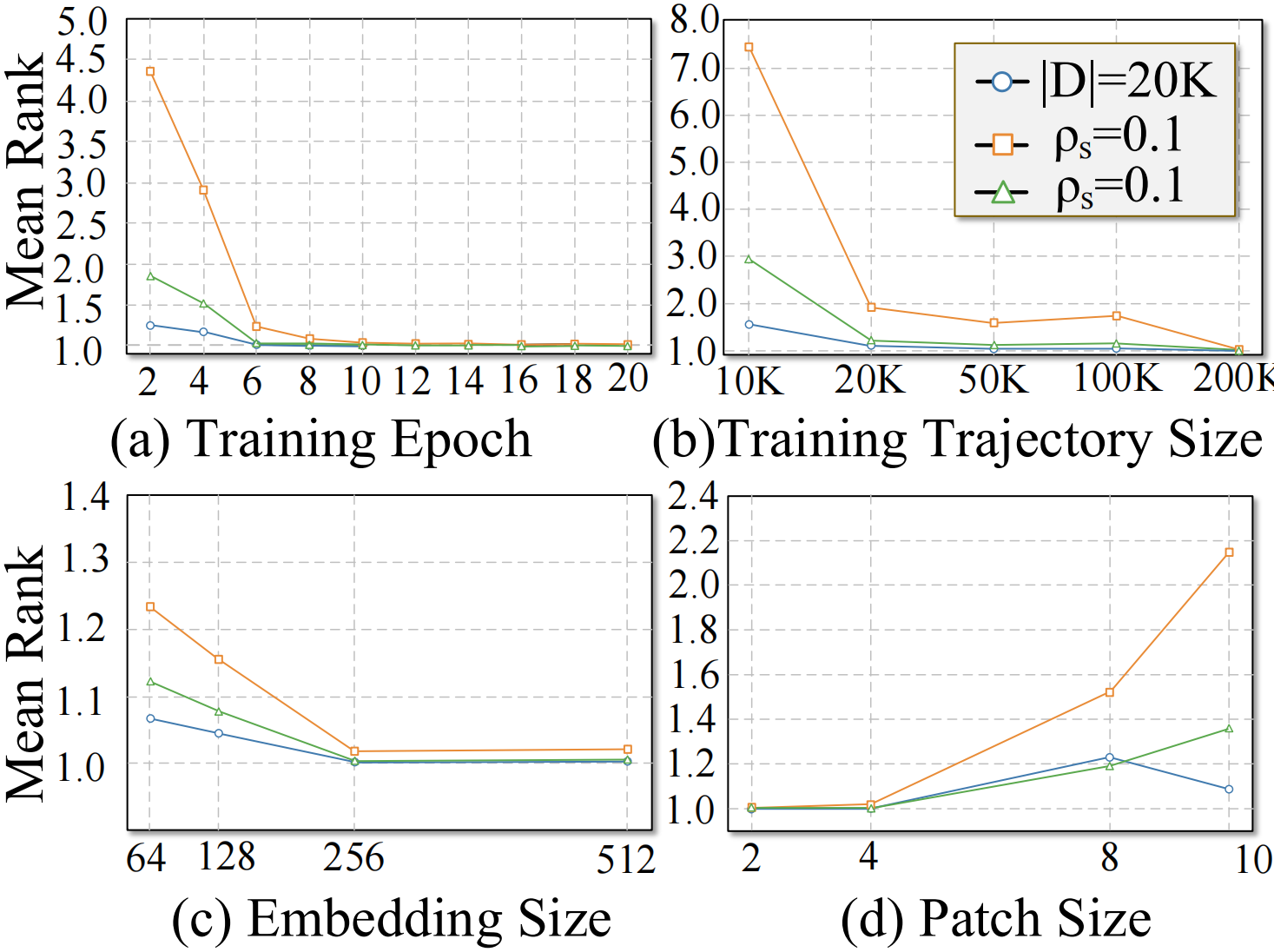}
\caption{Hyperparameter study (RQ6).}
    \label{fig:hyper}
\end{figure}

\paragraph{Training Epoch} Figure~\ref{fig:hyper}(a) shows that \movsem{} converges rapidly within 10 epochs, with stable performance extending to 20 epochs without overfitting. This rapid convergence reduces training costs while ensuring reliability.

\paragraph{Training Trajectory Size} Figure~\ref{fig:hyper}(b) shows that performance gains plateau around 20K trajectories for standard conditions, with degraded conditions requiring additional data for optimal performance. This provides practical guidance for minimum training data requirements.

\paragraph{Embedding Dimension} Figure~\ref{fig:hyper}(c) reveals optimal performance at dimensionalities in the range 256--512, with substantial improvement from smaller dimensions to 256, followed by stabilization. This indicates that the hierarchical encoding efficiently utilizes the embedding space without excessive parameters.

\paragraph{Patch Size} Figure~\ref{fig:hyper}(d) shows that patch size 4 achieves optimal performance. Smaller patches lack sufficient context, while larger patches dilute movement-semantics.

\section{Related Work}
\label{sec:related}

\paragraph{Traditional Trajectory Similarity Computation} Traditional methods rely on geometric and statistical principles. Alignment-based approaches adapt string matching algorithms, such as EDR~\cite{chen2005robust} which extends edit distance with spatial thresholds. Geometric methods include Hausdorff distance~\cite{hausdorff1914grundzuge}, measuring maximum point-to-nearest-neighbor distance, Fréchet distance~\cite{frechet1906quelques}, considering spatio-temporal ordering, and EDwP~\cite{ranu2015indexing} projecting trajectories onto a road network. However, these methods are computationally expensive and ignore movement semantics.

\paragraph{Learning-Based Trajectory Representation} Deep learning enables vector representations of trajectories that capture complex spatial-temporal patterns. Early approaches like t2vec~\cite{li2018deep} use sequence-to-sequence autoencoders, while TrjSR~\cite{cao2021accurate} and E2DTC~\cite{fang20212} employ recurrent architectures with attention mechanisms. Recent contrastive learning methods include TrajCL~\cite{chang2023trajcl} with trajectory-specific augmentations and dual-feature attention, and CLEAR~\cite{li2024clear} with multi-positive contrastive learning.
However, existing methods struggle with hierarchical movement modeling, computational efficiency for trajectories with many locations, and semantically-aware augmentations—limitations that \movsem{} addresses. 

\section{Conclusion}
\label{sec:conclusion}
We present \movsem{}, a movement-semantics contrastive learning framework that enriches GPS trajectories with movement-semantics features and uses hierarchical patch-based encoding with curvature-guided augmentation. Experiments show that \movsem{} is capable of outperforming state-of-the-art methods, reporting mean ranks close to 1 in similarity search, up to 20.3\% improvements in heuristic approximation, and up to 43.4\% faster inference, while maintaining the best robustness to data degradation.


\section{Acknowledgments}
This work was supported by Independent Research Fund Denmark (No. 1032-00481B).
\bigskip

\bibliography{aaai2026}

\clearpage

\appendix

\setcounter{secnumdepth}{2} 
{\centering\Large\bfseries Appendix\par}

\section{Curvature-Guided Augmentation}

The Curvature-Guided Augmentation (CGA) strategy is a core component of \movsem{} that generates physically plausible trajectory views for contrastive learning. Unlike naive random or block masking approaches that can create spatial discontinuities, CGA preserves behaviorally informative segments (such as turns and intersections) while selectively masking redundant straight-line segments.

\begin{algorithm}[htbp] 
\caption{Curvature‑Guided Augmentation (CGA)} 
\label{alg:cga} 
\begin{algorithmic}[1] 
\Require Trajectory $\mathcal{T}=\{(\bar{x}_i,\bar{y}_i)\}_{i=0}^{L-1}$, mask ratio $r_{\text{mask}}$, weights $(w_{\text{endpoint}},w_{\text{base}},w_{\text{direction}})$ 
\Ensure  Mask index set $\mathcal{M}$ 
\State $\mathcal{A}\leftarrow[]$ 
\For{$i=1$ \textbf{to} $L-2$} \Comment{Compute turning angles}
    \State $\vec v_{i-1}\!\gets\!(\bar{x}_i-\bar{x}_{i-1},\bar{y}_i-\bar{y}_{i-1})$, $\vec v_i\!\gets\!(\bar{x}_{i+1}-\bar{x}_i,\bar{y}_{i+1}-\bar{y}_i)$ 
    \State $\alpha_i\!\gets\!\arccos\!\big(\text{clip}(\langle\vec v_{i-1},\vec v_i\rangle/(\|\vec v_{i-1}\|\|\vec v_i\|), -1, 1)\big)$ 
    \State $\mathcal{A}.\text{append}(\alpha_i)$ 
\EndFor 
\State $\hat{\mathcal{A}}\gets[\alpha_i/\max(\mathcal{A}) \text{ for } \alpha_i \text{ in } \mathcal{A}]$ if $\max(\mathcal{A}) > 0$ else $[0, \cdots, 0]$ 
\For{$i=0$ \textbf{to} $L-1$} \Comment{Compute retention weights}
    \If{$i\in\{0,L-1\}$} 
        \State $w_i\gets w_{\text{endpoint}}$ \Comment{Preserve endpoints}
    \ElsIf{$i-1 < |\hat{\mathcal{A}}|$} 
        \State $w_i\gets w_{\text{base}}+\hat{\alpha}_{i-1}\cdot w_{\text{direction}}$ \Comment{Higher weight for turns}
    \Else 
        \State $w_i\gets w_{\text{base}}$ 
    \EndIf 
\EndFor 
\State $p_i\gets w_i/\sum_{j=0}^{L-1}w_j$ \Comment{Normalize to probabilities}
\State $\mathcal{M}\gets\text{MultinomialSample}(\{1-p_i\},\lfloor L\,r_{\text{mask}}\rfloor)$ 
\State \Return $\mathcal{M}$ 
\end{algorithmic} 
\end{algorithm}

\subsection{Algorithm Analysis}
The CGA algorithm operates in three main phases to generate semantically meaningful trajectory augmentations:

\textbf{Phase 1: Turning Angle Computation (Lines 2-6).} The algorithm first computes local turning angles for each interior point by calculating the angle between consecutive movement vectors. For each point $i$, it constructs two vectors: $\vec{v}_{i-1}$ representing movement from point $i-1$ to $i$, and $\vec{v}_i$ representing movement from point $i$ to $i+1$. The turning angle $\alpha_i$ is computed as the arccosine of the normalized dot product, with clipping to handle numerical precision issues. These angles capture local curvature information, with larger angles indicating sharper turns.

\textbf{Phase 2: Importance Weight Assignment (Lines 7-14).} The computed turning angles are normalized to the range $[0,1]$ to create comparable importance scores across different trajectories. The algorithm then assigns retention weights to each point based on three criteria: (1) \emph{Endpoint preservation}: Start and end points receive high weights ($w_{\text{endpoint}}$) to maintain trajectory boundaries; (2) \emph{Curvature-based weighting}: Interior points receive weights proportional to their normalized turning angles, with sharper turns getting higher retention probability; (3) \emph{Base weighting}: All points receive a minimum base weight ($w_{\text{base}}$) to ensure some straight segments are preserved for trajectory continuity.

\textbf{Phase 3: Probabilistic Sampling (Lines 15-16).} The weights are normalized into retention probabilities, and multinomial sampling is used to select which points to mask. The algorithm samples $\lfloor L \cdot r_{\text{mask}} \rfloor$ points for masking based on their inverse retention probabilities (i.e., points with lower retention weights have higher masking probability).

\subsection{Algorithm Properties and Complexity}
The CGA algorithm exhibits several desirable properties for trajectory augmentation:

\begin{itemize}[noitemsep, topsep=0pt]
    \item \textbf{Physical Plausibility}: By preserving high-curvature regions and endpoints, the algorithm generates trajectories that maintain realistic movement patterns without spatial jumps.
    
    \item \textbf{Semantic Awareness}: The CGA ensures that behaviorally informative segments (turns, intersections) are preferentially retained, while redundant straight-line segments are more likely to be masked.
    
    \item \textbf{Computational Efficiency}: The algorithm runs in linear time $O(L)$ with respect to trajectory length. Specifically: turning angle computation requires $O(L)$ time for a single pass through interior points; weight assignment processes each point once in $O(L)$ operations; and probabilistic sampling can be implemented in $O(L)$ time using efficient sampling algorithms.
    
    \item \textbf{Controllable Diversity}: The mask ratio $r_{\text{mask}}$ and weight parameters $(w_{\text{endpoint}}, w_{\text{base}}, w_{\text{direction}})$ provide fine-grained control over augmentation strength and characteristics.
\end{itemize}

The resulting augmented views are diverse yet semantically faithful, providing effective positive pairs for contrastive learning that focus on movement semantics rather than superficial coordinate patterns.

\subsection{Illustrative Examples}

\begin{figure}[h]
    \centering
    \includegraphics[width=\linewidth]{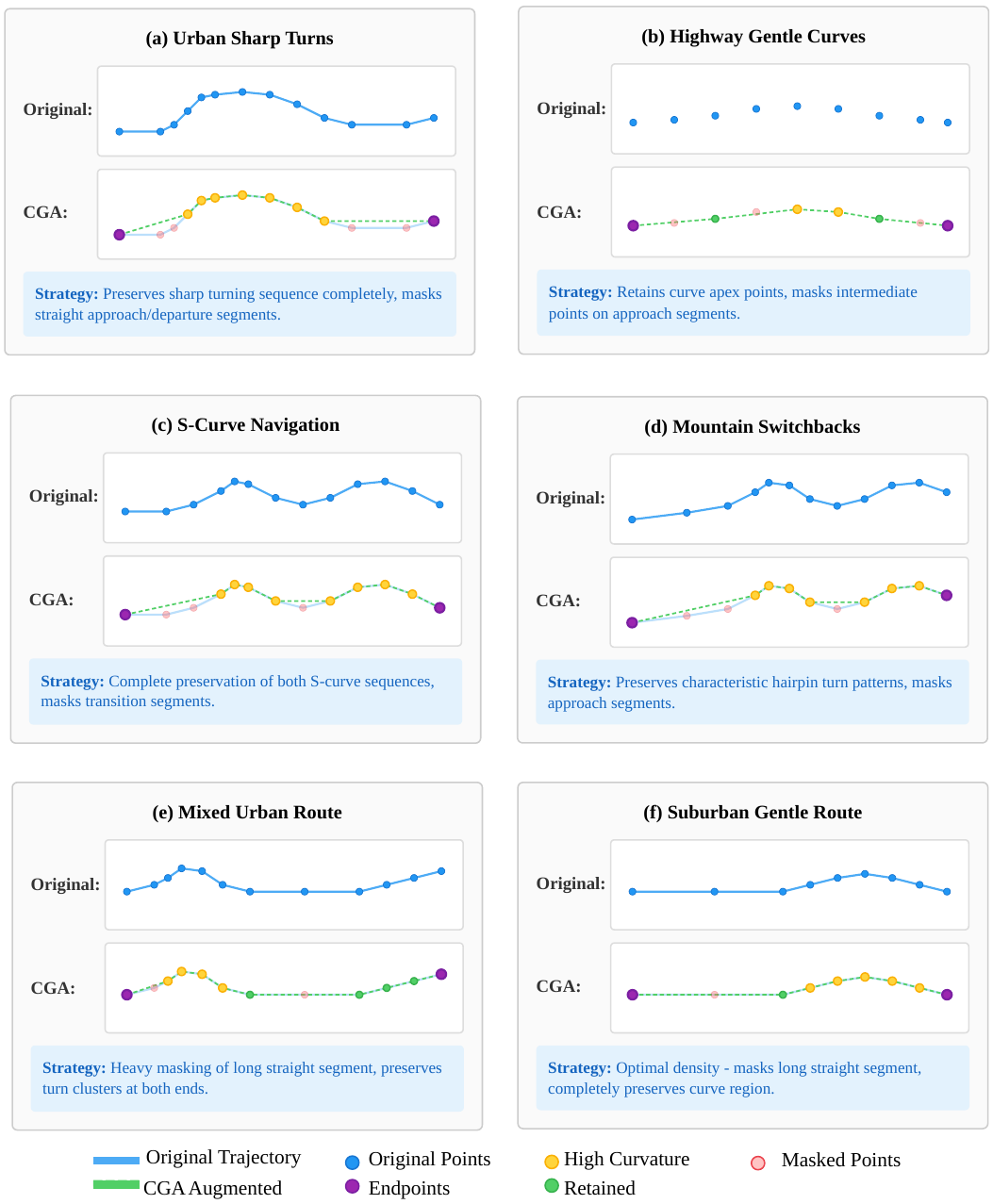}
\caption{Curvature-Guided Augmentation (CGA) Examples. Six trajectory masking scenarios showing CGA's semantic-aware strategy. Purple=endpoints (always preserved), yellow=high-curvature points (preferentially retained), green=retained points, red=masked points. (a) Urban turns, (b) highway curves, (c) S-curves, (d) switchbacks, (e) mixed routes, (f) suburban roads. CGA preserves movement semantics while generating physically plausible augmented views.}
    \label{fig:cga}
\end{figure}

Figure~\ref{fig:cga} demonstrates CGA's effectiveness across six representative trajectory scenarios: (a) urban sharp turns where complete turning sequences are preserved, (b) highway gentle curves where apex points are retained, (c) S-curve navigation with both curve sequences preserved, (d) mountain switchbacks maintaining hairpin patterns, (e) mixed urban routes with heavy masking of straight segments, and (f) suburban roads optimizing information density. These examples showcase how CGA consistently preserves high-curvature regions (yellow points) and endpoints (purple points) while selectively masking redundant straight segments (red points).

\section{Computational Complexity Analysis}


\subsection{Movement-Semantics Encoding Complexity}

The Movement-Semantics Encoding module operates with linear computational complexity $O(L)$ where $L$ is the trajectory length. The complexity breakdown is as follows:

\begin{itemize}[noitemsep, topsep=0pt]
    \item \textbf{Coordinate Normalization}: Mercator projection and region-based normalization require $O(L)$ operations for $L$ trajectory points.
    
    \item \textbf{Movement Dynamics Computation}: Computing displacement vectors $(dx_i, dy_i)$ and heading angles $\theta_i$ involves simple arithmetic operations for each point, scaling linearly as $O(L)$.
    
    \item \textbf{Spatial Graph Construction}: The trajectory-induced spatial graph has a one-time preprocessing cost of $O(|\mathcal{D}| \cdot \bar{L})$ where $\bar{L}$ is the average trajectory length across the dataset $\mathcal{D}$. However, the Node2Vec embedding lookup for each point during inference is $O(1)$ per point, contributing $O(L)$ to the overall complexity.
    
    \item \textbf{Feature Composition}: Concatenating pre-computed features requires $O(L)$ operations.
\end{itemize}

The overall complexity of Movement-Semantics Encoding is $O(L)$, establishing an efficient foundation for subsequent processing stages.

\subsection{Hierarchical Semantics Encoding Complexity}

The hierarchical encoding represents the core computational innovation of \movsem{}, achieving significant complexity reduction compared to traditional approaches.

\paragraph{Traditional Self-Attention Complexity}
Conventional Transformer-based trajectory methods~\cite{chang2023trajcl,xu2020trajectory} process the entire trajectory as a flat sequence of length $L$. The self-attention mechanism computes pairwise attention weights between all points, resulting in:

\begin{itemize}[noitemsep, topsep=0pt]
    \item \textbf{Attention Weight Computation}: $O(L^2 \cdot d_h)$ for computing the attention matrix
    \item \textbf{Linear Transformations}: $O(L \cdot d_h^2)$ for query, key, and value projections
    \item \textbf{Overall Complexity}: $O(L^2 \cdot d_h + L \cdot d_h^2)$, dominated by the quadratic term $O(L^2 \cdot d_h)$
\end{itemize}

For trajectories with hundreds of points, this quadratic scaling becomes computationally prohibitive and forces lossy downsampling that compromises movement fidelity.

\paragraph{Hierarchical Patch-Based Complexity}
Our hierarchical approach partitions trajectories into $M = \lceil L/P \rceil$ patches of size $P$, processing them through dual-level attention:

\begin{itemize}[noitemsep, topsep=0pt]
    \item \textbf{Intra-Patch Attention}: Each patch is processed independently with complexity $O(P^2 \cdot d_h)$. With $M$ patches, the total intra-patch complexity is:
    \begin{equation}
    O(M \cdot P^2 \cdot d_h) = O\left(\frac{L}{P} \cdot P^2 \cdot d_h\right) = O(L \cdot P \cdot d_h)
    \end{equation}
    
    \item \textbf{Inter-Patch Attention}: Operating on $M$ patch embeddings with complexity:
    \begin{equation}
    O(M^2 \cdot d_h) = O\left(\left(\frac{L}{P}\right)^2 \cdot d_h\right) = O\left(\frac{L^2}{P^2} \cdot d_h\right)
    \end{equation}
    
    \item \textbf{Total Hierarchical Complexity}:
    \begin{equation}
    O\left(L \cdot P \cdot d_h + \frac{L^2}{P^2} \cdot d_h\right)
    \end{equation}
\end{itemize}

For typical configurations where $P = 4$ and $L \gg P$, the complexity reduces to approximately $O(L \cdot d_h)$ since the linear term $L \cdot P \cdot d_h$ dominates. This represents a fundamental complexity reduction from $O(L^2 \cdot d_h)$ to $O(L \cdot d_h)$—achieving near-linear scaling compared to quadratic scaling of traditional approaches.

\paragraph{Complexity Comparison and Benefits}
Table~\ref{tab:complexity_comparison} summarizes the complexity comparison:

\begin{table}[ht]
\centering
\resizebox{\columnwidth}{!}{%
\begin{tabular}{lcc}
\toprule
\textbf{Method} & \textbf{Attention Complexity} & \textbf{Scaling} \\
\midrule
Traditional Self-Attention & $O(L^2 \cdot d_h)$ & Quadratic \\
Hierarchical (Ours) & $O(L \cdot d_h)$ & Linear \\
\textbf{Improvement} & \textbf{$L/P$ $\times$ faster} & \textbf{Linear vs. Quadratic} \\
\bottomrule
\end{tabular}}
\caption{Computational Complexity Comparison}
\label{tab:complexity_comparison}

\end{table}

This complexity reduction enables efficient processing of long trajectories without requiring lossy downsampling, preserving movement fidelity while achieving superior computational efficiency. The linear scaling is particularly beneficial for real-world applications where trajectories can contain hundreds of points.

\subsection{Overall System Complexity}

Building upon the complexity analysis of individual components, the complete \movsem{} framework achieves efficient complexity characteristics:

\begin{itemize}[noitemsep, topsep=0pt]
    \item \textbf{Training Complexity}: $O(L \cdot d_h)$ per trajectory for the forward pass, dominated by the hierarchical encoding
    \item \textbf{Inference Complexity}: $O(L \cdot d_h)$ per trajectory, enabling real-time similarity computation
    \item \textbf{Memory Complexity}: $O(L \cdot d_h + M \cdot d_h)$ where $M = \lceil L/P \rceil$, scaling linearly with trajectory length
    \item \textbf{Augmentation Overhead}: $O(L)$ per augmented view, ensuring minimal computational impact
\end{itemize}

This linear scaling across all components represents a significant improvement over existing quadratic methods, enabling \movsem{} to handle long trajectories efficiently while maintaining superior accuracy and semantic preservation. The efficient complexity profile makes \movsem{} suitable for real-time applications.

\section{Baselines and Datasets}


\subsection{Baselines}

\paragraph{Heuristic Methods (for RQ1--RQ3)}
Traditional geometric and statistical approaches form the foundation of trajectory similarity computation:

\begin{itemize}[noitemsep, topsep=0pt]
    \item \textbf{EDR}~\cite{chen2005robust}: Edit Distance with Real sequence that extends string edit distance to handle spatial trajectories with distance thresholds for point matching.
    
    \item \textbf{EDwP}~\cite{ranu2015indexing}: Edit Distance with Projections that projects trajectories onto road networks before computing edit distance to handle map-matched trajectories.
    
    \item \textbf{Hausdorff}~\cite{hausdorff1914grundzuge}: Measures the maximum distance from any point in one trajectory to the nearest point in another, capturing shape similarity but ignoring temporal ordering.
    
    \item \textbf{Fréchet}~\cite{frechet1906quelques}: Computes the minimum leash length needed to connect corresponding points when traversing both trajectories simultaneously, preserving temporal ordering.
    
    \item \textbf{CSTRM}~\cite{liu2022cstrm}: Canonical Stick Tensor Representation Method that represents trajectories as stick tensors and computes similarity via tensor operations.
\end{itemize}

\paragraph{Learning-Based Methods (for RQ1--RQ3)}
Recent deep learning approaches that learn trajectory representations:

\begin{itemize}[noitemsep, topsep=0pt]
    \item \textbf{t2vec}~\cite{li2018deep}: An early sequence-to-sequence autoencoder that learns trajectory embeddings by reconstructing spatial sequences with LSTM-based encoder-decoder architecture.
    
    \item \textbf{TrjSR}~\cite{cao2021accurate}: Trajectory Similarity Representation that combines recurrent neural networks with attention mechanisms to capture both local and global trajectory patterns.
    
    \item \textbf{E2DTC}~\cite{fang20212}: Enhanced Encoder for Deep Trajectory Clustering that uses bidirectional LSTM with attention for trajectory representation learning and clustering.
    
    \item \textbf{CLEAR}~\cite{li2024clear}: Contrastive Learning Enhanced trAjectory Representation that employs multi-positive contrastive learning with trajectory-specific data augmentations.
    
    \item \textbf{TrajCL}~\cite{chang2023trajcl}: A contrastive learning framework that uses dual-feature attention and trajectory-specific augmentations including masking, distortion, and sub-sampling strategies.
\end{itemize}

\paragraph{Supervised Methods (for RQ3)}
For heuristic similarity approximation experiments, we additionally compare with supervised methods designed for distance prediction:

\begin{itemize}[noitemsep, topsep=0pt]
    \item \textbf{TrajSimVec}~\cite{zhang2020trajectory}: A supervised method that learns trajectory representations using pairwise distance supervision.
    
    \item \textbf{TrajGAT}~\cite{yao2022trajgat}: Trajectory Graph Attention Network that models trajectories as graphs and uses graph attention mechanisms for similarity learning.
    
    \item \textbf{T3S}~\cite{yang2021t3s}: Transformer-based Trajectory-to-Trajectory Similarity learning that captures long-range dependencies in trajectory sequences.
\end{itemize}

\subsection{Dataset Characteristics}

\begin{table}[h]
\centering
\label{tab:dataset_stats}
\begin{tabular}{lrr}
\toprule
\textbf{Metric} & \textbf{Porto} & \textbf{Germany} \\
\midrule
Trajectories & 1,372,725 &  420,074 \\
Avg. Points & 48 & 72 \\
Max. Points & 200 & 200 \\
Avg. Length (km) & 6.37 & 252.49 \\
Max. Length (km) & 80.61 & 115,740.67 \\
\bottomrule
\end{tabular}
\caption{Dataset Statistics}

\end{table}
We evaluate on two complementary real-world trajectory datasets that represent different mobility patterns and scales:

\begin{itemize}[noitemsep, topsep=0pt]
    \item \textbf{Porto}: Dense urban taxi trajectories collected in Porto, Portugal from July 2013 to June 2014, representing short-distance urban mobility with frequent stops and turns. This dataset captures fine-grained urban movement patterns with complex intersection behaviors.
    
    \item \textbf{Germany}: Sparse long-distance trajectories collected across Germany from 2006 to 2013, capturing diverse inter-city travel patterns with longer segments and highway travel. This dataset represents coarse-grained mobility with emphasis on highway routing and long-distance connectivity.
\end{itemize}


The contrasting characteristics of these datasets—dense urban vs. sparse highway, short vs. long distances—enable comprehensive evaluation of \movsem{}'s effectiveness across diverse mobility scenarios.

\end{document}